\documentclass[final,5p,times,twocolumn]{elsarticle}

\usepackage{graphicx}
\usepackage{amsmath}
\usepackage{amssymb}
\usepackage[table,dvipsnames]{xcolor}
\usepackage{booktabs}
\usepackage{ulem}

\usepackage[pagebackref,breaklinks,colorlinks]{hyperref}
\usepackage{multirow}
\usepackage{colortbl}
\usepackage{float}
\usepackage{multicol}

\usepackage{siunitx}

\renewcommand{\emph}[1]{\textit{#1}}

\usepackage[capitalize]{cleveref}
\crefname{section}{Sec.}{Secs.}
\Crefname{section}{Section}{Sections}
\Crefname{table}{Table}{Tables}
\crefname{table}{Tab.}{Tabs.}

\begin{document}

\begin{frontmatter}



\title{SpectralGaussians: Semantic, spectral 3D Gaussian splatting for multi-spectral scene representation, visualization and analysis}





\author[1]{Saptarshi Neil Sinha}



\ead{saptarshi.neil.sinha@igd.fraunhofer.de}





\author[1]{Holger Graf}
\ead{holger.graf@igd.fraunhofer.de}


\author[2]{Michael Weinmann}


\ead{M.Weinmann@tudelft.nl}

\affiliation[1]{organization={Fraunhofer IGD},
    addressline={Fraunhoferstr. 5}, 
    city={Darmstadt},
    postcode={64283}, 
    country={Germany}}

\affiliation[2]{organization={Delft University of Technology},
    addressline={Van Mourik Broekmanweg 6}, 
    city={Delft},
    postcode={2628 XE}, 
    country={Netherlands}}

\begin{abstract}
We propose a novel cross-spectral rendering framework based on 3D Gaussian Splatting (3DGS) that generates realistic and semantically meaningful splats from registered multi-view spectrum and segmentation maps. This extension enhances the representation of scenes with multiple spectra, providing insights into the underlying materials and segmentation. We introduce an improved physically-based rendering approach for Gaussian splats, estimating reflectance and lights per spectra, thereby enhancing accuracy and realism.
In a comprehensive quantitative and qualitative evaluation, we demonstrate the superior performance of our approach with respect to other recent learning-based spectral scene representation approaches (i.e., XNeRF and SpectralNeRF) as well as other non-spectral state-of-the-art learning-based approaches.
Our work also demonstrates the potential of spectral scene understanding for precise scene editing techniques like style transfer, inpainting, and removal. Thereby, our contributions address challenges in multi-spectral scene representation, rendering, and editing, offering new possibilities for diverse applications.
%
\end{abstract}

\end{frontmatter}

\section{Introduction}
\label{sec:intro}

Accurate scene representation is an essential prerequisite for numerous applications. The way we perceive our surroundings in terms of a mixture of light gives us a particular scene understanding, thereby determining how we interact with our environment.
However, representing scenes in terms of red, green and blue color channels suffers from both a bad reproduction of the scene's appearance due to metamerism effects and lacking characteristics only observable in certain of the spectral bands.
Therefore, multi-spectral scene capture and representation has become of high relevance, where light and reflectance spectra are given with a higher resolution thereby surpassing the limitations of the broad-band RGB color model.
In domains such as architecture, automotive industries, advertisement, and design, accurate modeling of light transport and considering the full spectrum of light is crucial for virtual prototyping. Predictive rendering, which involves simulating the spectral transport of light, is necessary to assess and evaluate the visual quality of products before physical production. This ensures reliable assessment and enables color-correct scene reproduction.
Furthermore, spectral information as captured by multi-spectral (MS) cameras~\cite{RedEdge, Silios}, infrared (IR) cameras~\cite{infra-red}, and UV sensors~ \cite{ultra-violet} extends scene understanding in terms of insights on underlying material characteristics and behavior (including anomalies, defects, etc.) revealed only in certain sub-ranges of the light spectrum which empowers experts and autonomous systems to gain valuable insights and make informed decisions in the respective scenarios.
For \emph{precision} farming applications, multispectral scene monitoring enables
early detection and monitoring of harmful algal bloom in bodies of water~ \cite{algal_bloom}, facilitates the detection and classification of plant diseases~\cite{plant_disease, gray_mould_strawberry} to allow farmers to maintain crop health, optimize agricultural practices, and conduct quantitative and qualitative analysis of agro products~\cite{reviewfoodquality}, and allows getting insights on precise and objective plant parameters through 3D vision and multi-spectral imaging via phenotyping sensors like PlantEye~\cite{PlantEyeF600}.
In the context of cultural heritage, multi-spectral information is essential for gaining insights on production processes of artifacts or artworks and used materials, as e.g. relevant for the analysis of historical paintings~\cite{CH_intro_1, CH_intro_2, CH_intro_3} or for revealing hidden or altered features withing documents~\cite{document_analysis}, thereby also providing crucial hints on restoration of eroded parts by utilizing information from individual spectral bands that may exceed the visible range.
Among the many further application scenarios where multi-spectral scene monitoring and representation also allows for a more comprehensive understanding are facial recognition systems~\cite{face_recognition}, medical sciences, forensic sciences and remote sensing~\cite{application_all_areas}, where land cover and usage can be monitored more accurately.
Depending on the respective scenario, multi-spectral information can be either stored in terms of multi-channel representations~\cite{weinmann2017geospatial,zhan2017semisupervised,chen2018residual,chen2018igarss,palos2019mapping,weinmann2019jurse,sun2020weighted,shang2020multi,zhang2020mask,du2021incorporating,senchuri2021machine,florath2022glacier}
(as typically used for airborne or satellite-based surveillance), in terms of multi-spectral surface reflectance characteristics directly parameterized on 3D point clouds~\cite{weinmann2019fusion,mitschke2022hyperspectral,afifi2023tinto,rizaldy2023WHISPERS,rizaldy2024improving,rizaldy2024channel} or meshes~\cite{merzbach2017high,koutsoudis2021multispectral}, or in a volumetric manner as investigated with recent learning-based neural radiance field (NERF) representations~\cite{mildenhall2020nerf}.
Implicit scene representation using NeRFs~\cite{mildenhall2020nerf} has been demonstrated to allow high-fidelity scene representation based on training a neural network to predict view-dependent color and view-independent density information for points in the scene volume and leveraging volume rendering to predict the scene's appearance for particular viewpoints, while optimizing the network to produce images that match the original input images.
%
%
Beside the many extensions towards spatial representations, recent NeRF approaches also have explored extensions towards spectral scene representations~\cite{poggi2022xnerf,spectralnerf}. XNeRF and SpectralNeRF, despite their advancements in handling spectral scene representations, have limitations. XNeRF and SpectralNerf do not include reflectance and lighting estimation, segmentation of the spectral scene, and explicit geometry. These limitations can impact the accuracy, relightability, and comprehensive understanding of spectral scenes. Moreover, The 3DGS employs rasterization for rendering, which allows for real-time performance compared to NeRF-based methods and advanced 3DGS methods \cite{gaussian_grouping,qin2023langsplat} go beyond appearance and geometry modeling by supporting open-world and fine-grained scene understanding. They exceed the capabilities of NeRF-based approaches, like Semantic-NeRF \cite{zhi2021inplacescenelabellingunderstanding}, which incorporate semantic information into radiance fields for 3D scene modeling. However, these methods struggle to generalize to open-world scenarios. Distilled Feature Fields \cite{distilled_feature_fields_nerf} and LERF \cite{liu2023stylerf} explore distilling 2D features to aid in open-world 3D semantics, but they have limitations in accurate segmentation and cannot match the segmentation quality and efficiency of Gaussian-based methods \cite{gaussian_grouping,qin2023langsplat}.
%

%
The recently introduced 3D Gaussian Splatting (3DGS)~\cite{kerbl3Dgaussians} has been demonstrated to allow superior performance and quality compared to NERF-based scene representation and visualization.
This explicit scene representation replaces the neural network used in NeRF approaches with a set of Gaussians and the number and arrangement of Gaussians is optimized to best match the input data.
Thereby, the representation results in improved rendering efficiency, while also offering interpretability in contrast to black-box neural network representations.
However, the extension of 3DGS towards spectral scene representation and visualization has not been investigated so far.
In this paper, we present spectral 3D Gaussian splatting that allows efficient multi-spectral scene representation and visualization.
For this purpose, we present the following key contributions:
\begin{itemize}
  \item We present a novel cross-spectral rendering framework that extends the scene representation based on 3D Gaussian Splatting (3DGS) to generate realistic and semantically meaningful splats from registered multi-view spectrum and segmentation maps.
  \item We present an improved physically-based rendering approach for Gaussian splats, estimating reflectance and lights per spectra, which enhances the accuracy and realism of the rendered output by considering the unique characteristics of different spectra, resulting in visually convincing and physically accurate scene representations.
\item We generated two synthetic spectral datasets by extending the shiny Blender dataset~\cite{verbin2022refnerf} and the synthetic NERF dataset~\cite{mildenhall2020nerf} in terms of their spectral properties. The datasets were created through simulations using Mitsuba~\cite{Mitsuba3}, where scenes were rendered at various wavelengths across the visible spectrum. These datasets are expected to serve as valuable resources for researchers and practitioners, offering a diverse range of spectral scenes for experimentation, evaluation, and advancements in the field of image-based/multi-view spectral rendering. We plan to release both the dataset and the code to generate similar datasets using Mitsuba~\cite{Mitsuba3}, promoting reproducibility and further contributions to the field.
\item In the scope of a detailed evaluation on our datasets, as well as the spectral NeRF dataset ~\cite{spectralnerf}, we showcase the potential of our approach in spectral scene understanding. Through our evaluation, we demonstrate that spectral scene understanding enables efficient and accurate scene editing techniques, including style transfer, in-painting, and removal. These techniques leverage the specific spectral characteristics of objects in the scene, facilitating more precise and context-aware modifications.
\end{itemize}

\section{Related work}

\subsection{Learning-based scene representation}
In recent years, significant advancements have been made in generating photo-realistic novel views through the use of novel learning-based scene representations combined with volume rendering techniques.
Neural Radiance Fields (NeRF) \cite{mildenhall2020nerf,tewari2022advances} represent the scene based on a neural network that predicts local density and view-dependent color for points in the scene volume. This information can then be used to synthesize images of the scene using volume rendering techniques.
The network representing the scene is trained by minimizing the deviation of the predicted images to their respective given input images under the respective view conditions, thereby exploiting the observation that an accurate scene representation by the network leads to an accurate image synthesis.
The remarkable potential of the NeRF approach for novel view synthesis has given rise to several notable extensions. Researchers have focused on improving rendering quality by addressing issues such as aliasing~\cite{barron2021mipnerf, wang2022nerf, barron2022mipnerf360, barron2023zip}, as well as accelerating network training~\cite{reiser2021kilonerf, fridovich2022plenoxels, mueller2022instant, chen2022tensorf, Yariv2023baked}. Furthermore, there have been efforts to handle more complex inputs, including unconstrained image collections~\cite{martin2021nerf, chen2022hallucinated, Seong2022hdrplenoxels}, image collections requiring the refinement or complete estimation of camera pose parameters~\cite{wang2021nerf,yen2021inerf,lin2021barf,jeong2021self}, deformable scenes~\cite{park2021nerfies,pumarola2021d} and large-scale scenarios~\cite{tancik2022block,turki2022mega,Mi2023switchnerf}. Further works aimed at guiding the training and handling textureless regions by incorporating depth cues~\cite{wei2021nerfingmvs,deng2022depth,roessle2022dense,rematas2022urban,attal2021torf}.
Despite the great success of NeRFs for novel view synthesis applications, the neural network lacks interpretability and the extraction of surface information requires network evaluations on a dense grid and a subsequent derivation of surface information from the volumetric density information based on techniques like Marching Cubes~\cite{lorensen1998marching}, which limits real-time applications.
Therefore, further works focused on representing scenes in terms of implicit surfaces~\cite{wang2021neus, wang2023neus2, ge2023ref}, explicit representations using points \cite{xu2022point}, meshes \cite{munkberg2022extracting}, and 3D Gaussians \cite{kerbl3Dgaussians}.
Point-based neural rendering techniques, such as Point-NeRF~\cite{xu2022point}, merge precise view synthesis from NeRF with the fast scene reconstruction abilities of deep multi-view stereo methods. These techniques employ neural 3D point clouds to enable efficient rendering, thereby facilitating accelerated training processes.
Furthermore, a recent approach~\cite{zhang2023frequency} has shown that point-based methods are well-suited for scene editing purposes.
Recently, 3D Gaussian Splatting \cite{kerbl3Dgaussians} has been introduced as the state-of-the-art, learning-based scene representation based on optimized Gaussians for novel view synthesis, surpassing existing implicit neural representation methods such as NeRFs in terms of both quality and efficiency. This approach utilizes anisotropic 3D Gaussians as an explicit scene representation and employs a fast tile-based differentiable rasterizer for image rendering.
However, extending these novel scene representations to the spectral domain beyond RGB channels remains an open challenge, with only a few seminal works addressing this so far. Spectral variants of NERF, such as xNERF~\cite{poggi2022xnerf} for cross-spectral spectrum-maps and SpectralNeRF~\cite{spectralnerf} for multi-spectral spectrum-maps, have shown effectiveness in generating novel views across different spectral domains. 
The cross-spectral splats generated by our approach can be visualized via an interactive spectral viewer~\cite{spectralsplatsviewer2024interactive} based on Viser~\cite{nerfstudio}. Besides view synthesis, the viewer allows to visualize splats, even with spectral characteristics, as well as visualizing residuals between different versions of splats such as splats from different iterations during training or comparing differences between splats in different spectral ranges. Furthermore, the user study conducted in their work~\cite{spectralsplatsviewer2024interactive} validates the effectiveness and practicality of the reconstructed 3D splats derived from the spectrum maps, confirming their utility in spectral visualization and analysis. However, the framework of reconstructing a spectral Gaussian Splatting scene representation is a novel contribution in this paper and has not been considered in their work~\cite{spectralsplatsviewer2024interactive}.

\subsection{Radiance based appearance capture}

Instead of focusing on the pure reproduction of a scene according to the original NeRF formulation without explicitly modeling reflectance and illumination characteristics, several NeRF extensions focused on modeling reflectance by separating visual appearance into lighting and material properties. Respective approaches have the capability to jointly predict environmental illumination and surface reflectance properties even in the presence of unknown or varying lighting conditions\mbox{~\cite{bi2020neural, zhang2021physg, boss2021nerd, srinivasan2021nerv, boss2021neural, zhang2021nerfactor}}.
%
%

One notable contribution is Ref-NeRF ~\cite{verbin2022refnerf}, which introduces a novel parameterization and structuring of view-dependent outgoing radiance, along with a regularizer on normal vectors. This enhances the accuracy in predicting reflectance properties. To address the challenge of learning geometry from highly specular surfaces, recent works ~\cite{liu2023nero, liang2023envidr, liang2022spidr} have utilized SDF-based representations. This enables more precise estimation of surface normals for physically based rendering. However, these methods suffer from time-consuming optimization and slow rendering speed, limiting their practical application in real-world scenarios.
Furthermore, 
NVDiffRec \cite{hasselgren2022nvdiffrecmc} is an explicit representation method that directly optimizes triangle meshes with materials and environment map lighting, enabling real-time interactive applications, unlike MLP-based methods that tend to be slower. 
 
Relightable Gaussians \cite{R3DG2023} presents a differentiable point-based rendering framework for material and lighting decomposition from multi-view images, enabling real-time relighting and editing of 3D point clouds. It surpasses existing material estimation approaches and offers improved results. GaussianShader \cite{jiang2023gaussianshader} is another method that enhances neural rendering in scenes with reflective surfaces by applying a simplified shading function on 3D Gaussians. It addresses the challenge of accurate normal estimation on discrete 3D Gaussians, achieving a balance between efficiency and rendering quality. Our shading model is inspired by this method 
%
where we use the model without the residual color in the reflectance estimation.

\subsection{Sparse spectral scene understanding}

Gaussian splatting based semantic segmentation frameworks, such as Gaussian Grouping~\cite{gaussian_grouping} and LangSplat~\cite{qin2023langsplat}, have successfully utilized foundation models like Segment Anything~\cite{kirillov2023segany} to segment scenes. LangSplat is a 3D language field that enables precise and efficient open-vocabulary querying within 3D spaces by representing language features using a collection of 3D Gaussians distilled from CLIP~\cite{radford2021learning}. Gaussian Grouping extends Gaussian Splatting by incorporating object-level scene understanding and introducing Identity Encodings to reconstruct and segment objects in open-world 3D scenes. We utilized this method for accurate semantic segmentation of spectral scenes. Segmenting the scene per spectra provides valuable information about regions that are visible in specific spectral ranges, enabling us to obtain finer details that can be leveraged in various domains such as cultural heritage~\cite{CH_intro_1, CH_intro_2, CH_intro_3}, smart farming~\cite{algal_bloom, reviewfoodquality, gray_mould_strawberry}, document analysis~\cite{document_analysis}, face recognition~\cite{face_recognition}, and other fields. This spectral segmentation approach offers insights and solutions for diverse applications in these domains.
%
%
%
In the scope of the evaluation, we demonstrate that spectral scene understanding enables efficient and accurate scene editing techniques, including style transfer, in-painting, and removal.

\subsection{Spectral renderers}
Spectral rendering engines such as ART~\cite{ART}, PBRT v3~\cite{pharr2016physically}, and Mitsuba~\cite{mitsuba2} are commonly utilized by the scientific community. While CPU-based renderers are more prevalent, there is a growing trend of GPU-based spectral renderers that leverage GPU acceleration. Some examples of GPU-based spectral renderers include Mitsuba 2~\cite{mitsuba2-retargetable}, PBRT v4~\cite{pbrt-v4}, and Malia~\cite{malia}. These renderers play a crucial role in simulating real-world spectral data and are gaining recognition in the field. 
To achieve computational efficiency in deep learning and focus on relevant spectral information, we adopt a sparse spectral rendering approach using multi-view spectrum maps. This technique enables faster computations by reducing unimportant spectral data while preserving the necessary information for realistic rendering of spectral scenes. By leveraging spectrum maps from multiple viewpoints, high-quality spectral renderings are generated with a reduced computational cost compared to full-resolution spectral rendering methods.

\section{Background}
The human eye is sensitive to only 
%
a certain range in
the electromagnetic spectrum (for wavelengths between about 380nm and 780nm) which varies between subjects. The response curve of the human eye is to the red, green and blue wavelengths were determined using color matching functions which has been standardised by CIE in 1932 \cite{color_science}. Given a spectral power distribution \(L(\lambda)\), its corresponding CIE tristimulus values X, Y and Z can be computed by convolution of the \(L(\lambda)\) with the appropriate color matching functions \(f_{X}(\lambda)\),\(f_{Y}(\lambda)\), \(f_{Z}(\lambda)\) as represented in the following equations~\cite{star_spectral_rendering}:

\begin{equation}\label{eqn:1}
    \begin{cases}
     X = \int_{380}^{780} f_{X}(\lambda)L(\lambda)d\lambda\\
     Y = \int_{380}^{780} f_{Y}(\lambda)L(\lambda)d\lambda\\
     Z = \int_{380}^{780} f_{Z}(\lambda)L(\lambda)d\lambda\\
    \end{cases}       
\end{equation}

The spectral power distribution \(L(\lambda)\) at a point \(x\) for incoming wavelength \(\lambda_{i}\) and outgoing wavelength \(\lambda_{o}\) can be computed as follows:

\begin{equation}\label{eqn:2}
\resizebox{\linewidth}{!}{$
\begin{split}
L(x,\omega_{i}, \omega_{o}, \lambda_{i},\lambda_{o}) = \int_{\Omega} f_{r}(x,\omega_{i}, \omega_{o}, \lambda_{i}, \lambda_{o})L_{i}(x,\omega_{i}, \omega_{o}, \lambda_{i}) \cos\theta \, d\omega_{i}
\end{split}
$}
\end{equation}
where  \(\Omega\) represents the hemisphere above a surface point \(x\), \(f_{r}\) is the bidirectional reflectance function, \(L_{i}\) is the incoming radiance coming from incident direction \(\omega_{i}\) and \(\omega_{o}\) is the direction of the outgoing radiance.


The final RGB image is obtained based on the conversion from the XYZ color space to the sRGB space which involves the following steps. 
\begin{itemize}
    \item \textbf{Conversion to linear RGB}: This step involves using a matrix multiplication to convert XYZ values to linear RGB values.
    
     \begin{equation}\label{eqn:3}
            \begin{pmatrix} R \\ G \\ B \end{pmatrix} = 
            \begin{pmatrix}  M^{l} \end{pmatrix} 
            \begin{pmatrix} X \\ Y \\ Z \end{pmatrix}
    \end{equation}
    There are many methods \cite{smits2000} to convert XYZ to linear RGB and the value of the matrix \(M^{l}\) depends on it. 
    
    \item \textbf{Gamma correction}: Linear RGB values are gamma-corrected to get sRGB values. This involves applying a power function with a specific gamma value (\(\approx 2.2\)).

    \item \textbf{Clipping}: All RGB values are clipped within the range [0, 1].
\end{itemize}

The above steps can be combined to get the final transformation matrix (\(M^{c}\)) directly get the sRGB values:
\begin{equation}\label{eqn:4}
        \begin{pmatrix} R \\ G \\ B \end{pmatrix} = 
        \begin{pmatrix}  M^{c} \end{pmatrix} 
        \begin{pmatrix} X \\ Y \\ Z \end{pmatrix}
\end{equation}

Based on equations(\ref{eqn:1}), (\ref{eqn:2}), and (\ref{eqn:4}), the RGB values per spectra maps~\cite{spectralnerf} can be computed according to
\begin{equation}\label{eqn:5}
    \begin{pmatrix} R_{\lambda} \\ G_{\lambda} \\ B_{\lambda} \end{pmatrix} = 
\begin{pmatrix} 
    M^{c}_{11}f_X(\lambda) + M^{c}_{12}f_Y(\lambda) + M^{c}_{13}f_Z(\lambda) \\
    M^{c}_{21}f_X(\lambda) + M^{c}_{22}f_Y(\lambda) + M^{c}_{23}f_Z(\lambda) \\
    M^{c}_{31}f_X(\lambda) + M^{c}_{32}f_Y(\lambda) + M^{c}_{33}f_Z(\lambda) \\
\end{pmatrix} L(\lambda) \Delta\lambda
\end{equation}

\section{Methodology}

\subsection{Spectral Gaussian splatting}
\begin{figure*}[tb]
  \centering
  \includegraphics[width=\textwidth]{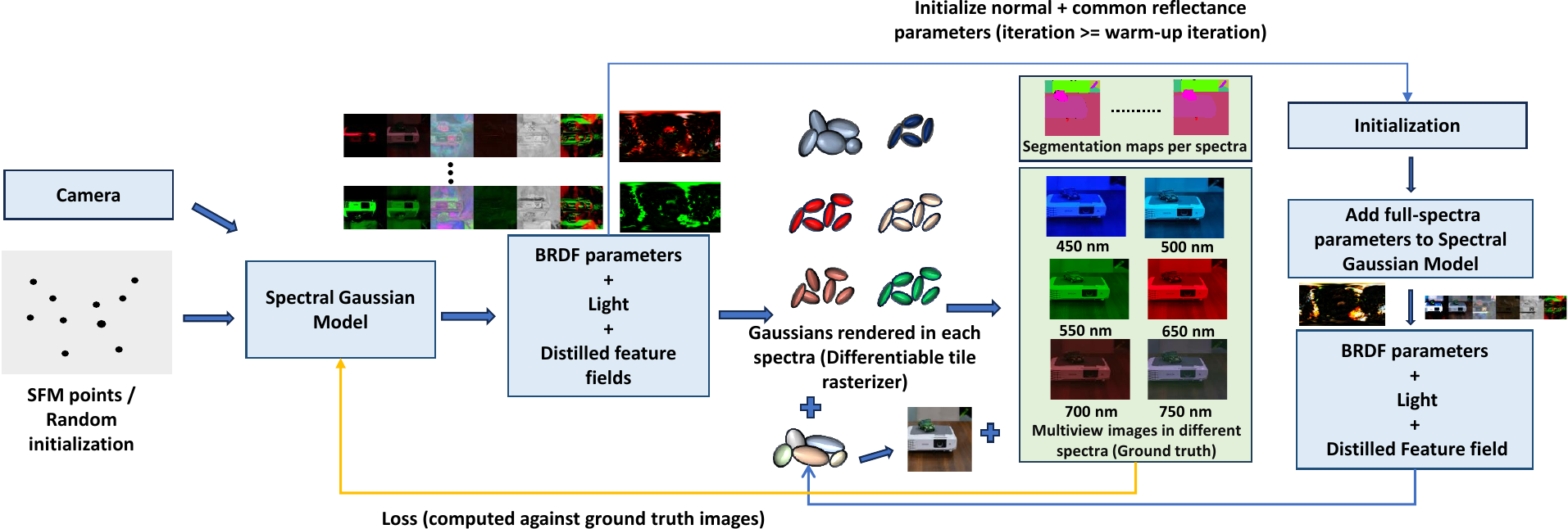}
  \caption{The proposed spectral Gaussian splatting framework: 
  %
  Spectral Gaussian model predicting BRDF parameters, distilled feature fields, and light per spectrum from multi-view spectrum-maps. The full-spectra maps and learnable parameters are introduced later in the training process by initializing them with priors from all other spectra.}
  \label{fig:pipeline}
\end{figure*}
We propose an end-to-end spectral Gaussian splatting approach that enables physically-based rendering, relighting, and semantic segmentation of a scene. Our method is built upon the Gaussian splatting architecture~\cite{kerbl3Dgaussians} and leverages the Gaussian shader~\cite{jiang2023gaussianshader} for the accurate estimation of BRDF parameters and illumination. By employing Gaussian grouping~\cite{gaussian_grouping}, we effectively group 3D  Gaussian splats with similar semantic information. Our framework excels in generating full spectra rendering and conveniently initializes common features from other spectra trained to a specific iteration, ensuring  improved reconstruction of splats.
%
In Figure~ \ref{fig:pipeline}, we showcase our proposed spectral Gaussian splatting framework, which uses a Spectral Gaussian model to predict BRDF parameters, distilled feature fields, and light per spectrum from multi-view spectrum-maps. Our method combines segmentation, appearance modeling, and sparse spectral scene representation in an end-to-end manner. Thereby it enhances BRDF estimation by incorporating spectral information. The framework has applications in material recognition, spectral analysis, reflectance estimation, segmentation, illumination correction, and inpainting.
%
%

%
In the following subsections, we provide further details regarding 
%
%
the spectral model, covering topics such as appearance modeling, spectral semantic scene representation, spectral scene editing, and the seamless integration of these aspects into the 3DGS framework.
%
\subsection{Spectral appearance modelling}
\label{sec:spec_appearance_modelling}
In order to support material editing and re-lighting, we use an enhanced representation of appearance by replacing the spherical  harmonic co-efficients by a shading function, which incorporates diffuse color, roughness, specular tint and normal information and a differentiable environment light map to model direct lighting similar to the Gaussian shader~\cite{jiang2023gaussianshader}

Thereby, the rendered color per spectrum of a Gaussian sphere can be computed by considering its diffuse color, specular tint, direct specular light, normal vector and roughness according to
\begin{equation}
\label{eq:appearance_modelling}
c(\omega_o)_{\lambda} = \gamma \left( c_{d_{\lambda}} + s_{\lambda} \odot L_{s_{\lambda}}(\omega_o, n, \rho_{\lambda}) \right)
\end{equation}
where,\(c(\omega_o)_{\lambda}\) represents the rendered color per spectrum for the viewing direction \(\omega_o\). The function \(\gamma\) is a gamma tone mapping function that adjusts the color values for display purposes. \(c_{d_{\lambda}} \in [0, 1]^3\) denotes the diffuse color of the Gaussian sphere, specifying the color appearance under diffuse lighting per spectrum. \(s_{\lambda} \in [0, 1]^3\) is the specular tint on the sphere, indicating the color of the specular highlights per spectrum. \(L_{s_{\lambda}}(\omega_o, n, \rho_{\lambda})\) describes the direct specular light for the Gaussian sphere in the viewing direction \(\omega_o\) per spectrum, considering the surface normal \(n\) and roughness \(\rho_{\lambda}\). \(n\) is the normal vector indicating the surface orientation, and \(\rho_{\lambda} \in [0, 1]\) represents the surface smoothness or roughness per spectrum.

The shading model is motivated by two aspects:
\begin{itemize}
    \item The diffuse color ($c_{d_{\lambda}}$) represents the consistent colors of the Gaussian sphere and remains unchanged with viewing directions.
    \item The term $s \odot L_{s_{\lambda}}(\omega_o, \mathbf{n}, \rho_{\lambda})$ describes the interaction between the intrinsic surface color $s_{\lambda}$ (specular tint) and the direct specular light $L_{s_{\lambda}}$. This term accounts for most of the reflections in rendering.
    %
\end{itemize}
To compute the specular light per spectrum $L_{s_{\lambda}}$ in the shading model, the incoming radiance is integrated with the specular GGX 
Normal Distribution Function $D$ \mbox{~\cite{Walter2007MicrofacetMF}}. The integral is taken over the entire upper semi-sphere $\Omega$ and is given by:
\begin{equation}
L_{s_{\lambda}}(\omega_o, \mathbf{n}, \rho_{\lambda}) = \int_{\Omega} L(\omega_i)D(\mathbf{r}, \rho_{\lambda})(\omega_i \cdot \mathbf{n}) d\omega_i
\end{equation}

Here, $\Omega$ represents the whole upper hemi-sphere, $\omega_i$ is the direction for the input radiance, and $D$ characterizes the specular lobe (effective integral range). The reflective direction $\mathbf{r}$ is calculated using the view direction $\omega_o$ and the surface normal $\mathbf{n}$ as $\mathbf{r} = 2(\omega_o \cdot \mathbf{n})$ \hspace{1px}$\mathbf{n} - \omega_o$.
%
%
\(L_{s_{\lambda}}\) represents the direct specular light per spectral band \(\lambda\).

\subsection{Spectral semantic scene representation}

%
Per-spectrum segmentation maps serve multiple purposes in various applications. They enable sparse scene representation, allowing for detailed identification of specific regions of interest and the detection of attributes like material composition or texture. These maps are beneficial for tasks like inpainting and statue restoration, where spectral information is crucial for accurate and realistic results. Additionally, per-spectrum segmentation maps aid in anomaly detection by analyzing the spectral properties of different regions and identifying deviations from expected patterns. This approach of segmenting different spectra enables the identification of specific regions of interest, such as the detection of grey mould disease in strawberry plants \cite{gray_mould_strawberry}. Overall, these maps provide valuable insights into the scene, allowing for more robust and precise image processing and analysis.
%
Our framework utilizes the Gaussian grouping method \cite{gaussian_grouping} to generate per-spectrum segmentation of the splats. This ensures consistent mask identities across different views of the scene and groups 3D Gaussian splats with the same semantic information. To create ground truth multi-view segmentation maps for each spectrum, we employ the Segment Anything Model (SAM) \cite{kirillov2023segany} along with a zero-shot tracker \cite{deva}. This combination automatically generates masks for each image in the multi-view collection per spectrum, ensuring that each 2D mask corresponds to a unique identity in the 3D scene. By associating masks of the same identity across different views, we can determine the total number of objects present in the 3D scene. 
In addition to the existing appearance and lighting properties, a novel attribute called \emph{Identity Encoding} is assigned to each spectral Gaussian, similar to Gaussian grouping \cite{gaussian_grouping}. The Identity Encoding is a compact and learnable vector (of length 16) that effectively distinguishes different objects or parts within the scene. During training, similar to using Spherical Harmonic coefficients to represent color, the method optimizes the Identity Encoding vector to represent the instance ID of the scene. Unlike view-dependent appearance modeling, the instance ID remains consistent across different rendering views, as only the direct-current component of the Identity Encoding is generated by setting the Spherical Harmonic degree to 0.
\\
The final rendered 2D mask identity feature, denoted as \(E_{id_{\lambda}}\), for each pixel per spectrum \(\lambda\) is calculated by taking a weighted sum over the Identity Encoding (\(e_{i_{\lambda}}\)) of each Gaussian per spectrum. The weights are determined by the influence factor \(\alpha'_{i_{\lambda}}\) of the respective Gaussian on that pixel per spectrum. Mathematically, this can be expressed as
\begin{equation}
\label{eq:identity_sh}
E_{id_{\lambda}} = \sum_{i \in N} e_{i_{\lambda}} \alpha'_{i_{\lambda}} \prod_{j=1}^{i-1} (1 - \alpha'_{j_{\lambda}})\
\end{equation}
where $N$ represents the total number of Gaussians.

To group the 3D Gaussians based on their object mask identities, a grouping loss \(L_{id_{\lambda}}\) is computed per spectra. This loss has two components 
%
%
, i.e. it can be formulated as
\begin{equation}
    L_{id_{\lambda}} = L_{2d_{\lambda}}  + L_{3d_{\lambda}} 
\label{eq:total_loss_seg}
\end{equation}
where the first component \(L_{2d_{\lambda}}\) is the 2D Identity Loss, which involves a softmax function to classify the rendered 2D features \(E_{id}\) (see Equation \ref{eq:identity_sh}) into \(K_{s}+1\) categories, representing the total number of masks per spectrum in the 3D scene. The standard cross-entropy loss \(\mathcal{L}
_{2d_{\lambda}}\) for the classification of \(K_{s}+1\) categories is applied. 
So given the rendered 2D features $E_{id_{\lambda}}$ as input, a linear layer is first applied $f$ to restore its feature dimension back to $K$:
\begin{equation}
 \textit{f}(E_{id_{\lambda}}) = W \cdot E_{id_{\lambda}} + b,
\end{equation}
where $W$ represents the learnable weight matrix and $b$ is the bias term.
To obtain the probabilities for each category, we apply the softmax function:
\begin{equation}
\text{softmax}(f(E_{id_{\lambda}})) = \frac{\exp(f(E_{id_{\lambda}}))}{\sum_{i=1}^{K} \exp(f(E_{id_{\lambda}}))},
\end{equation}
For the identity classification task with $K$ categories per spectrum \(\lambda\), we utilize the standard cross-entropy loss:
\begin{equation}
L_{2d_{\lambda}} = -\sum_{i=1}^{K} y_{i_{\lambda}} \log(\text{softmax}(f(E_{id_{\lambda}}))),
\end{equation}
where $y$ is the ground truth label for each category.
The second component is the 3D Regularization Loss $\mathcal{L}_{3d_s}$, which capitalizes on the 3D spatial consistency to regulate the learning process of the Identity Encoding \(e_{i_{\lambda}}\) per spectrum \(\lambda\). This loss ensures that the Identity Encodings of the top \(k\)-nearest 3D Gaussians are similar in terms of their feature distance, thereby promoting spatially consistent grouping. The 3D grouping loss per spectrum $\lambda$ and sampled $m$ points is computed as:
\begin{equation}
\scalebox{0.85}{$\displaystyle\mathcal{L}_{3d_{\lambda}} = \frac{1}{m} \sum_{j=1}^{m} D_{{kl}}(P || Q) = \frac{1}{mk} \sum_{j=1}^{m}\sum_{i=1}^{k} F(e_{j_{\lambda}}) \log\left(\frac{F(e_{j_{\lambda}})}{F(e'_{i_{\lambda}})}\right)$}
\end{equation}
Here, $P$ contains the sampled Identity Encoding $e_{\lambda}$ of a 3D Gaussian, and $Q = {e'_{1_{\lambda}}, e'_{2_{\lambda}}, ..., e'_{k_{\lambda}}}$ represents its $k$ nearest neighbors in 3D Euclidean space. 
%

\subsection{Combined (Semantic and appearance) spectral model}
Combined with the original 3D Gaussian loss 
\cite{kerbl3Dgaussians} (we use \(\gamma\) instead of \(\lambda\) as we use \(\lambda\) to denote the spectral bands) on image rendering (we use the appearance model as explained in the Sec.~\ref{sec:spec_appearance_modelling} instead of spherical harmonics), the total loss per spectra \(\mathcal{L}_{render_{s}}\) for fully end-to-end training is given by
\vspace{-0.2cm}
\begin{equation}
\label{eq:gaussiang+grouping_loss}
\scalebox{0.9}{$\displaystyle \mathcal{L}_{\text{render}_{\lambda}} =  (1 - \gamma)L_{1_{\lambda}} + \gamma \cdot \mathcal{L}_{\text{D-SSIM}_{\lambda}} + \gamma_{2d_{\lambda}}\mathcal{L}_{2d_{\lambda}} + \gamma_{3d}\mathcal{L}_{3d_{\lambda}}$}
\end{equation}
The total loss is given by
\begin{equation}
\mathcal{L}_{\text{render}_{total}} = \sum_{\lambda=1}^{n_{\lambda}}\mathcal{L}_{\text{render}_{\lambda}}
\end{equation}
 where $n_{\lambda}$ is the total number of spectral bands. 
 
To enhance the optimization process and improve robustness, the model is initially trained for a specific warm-up iteration (1000 iterations) without incorporating the full-spectra spectrum maps. Following this, the common BRDF parameters and normals for the full-spectra are initialized (see Fig.~\ref{fig:pipeline}) using the average values from all other spectra, and this initialization step is integrated into the training process. By including these adequate priors, the optimization of parameters is guided more effectively, leading to better outcomes as demonstrated in the quantitative and qualitative analysis.
 \begin{figure}[htb!]
  \centering
  \includegraphics[width=0.5\textwidth]{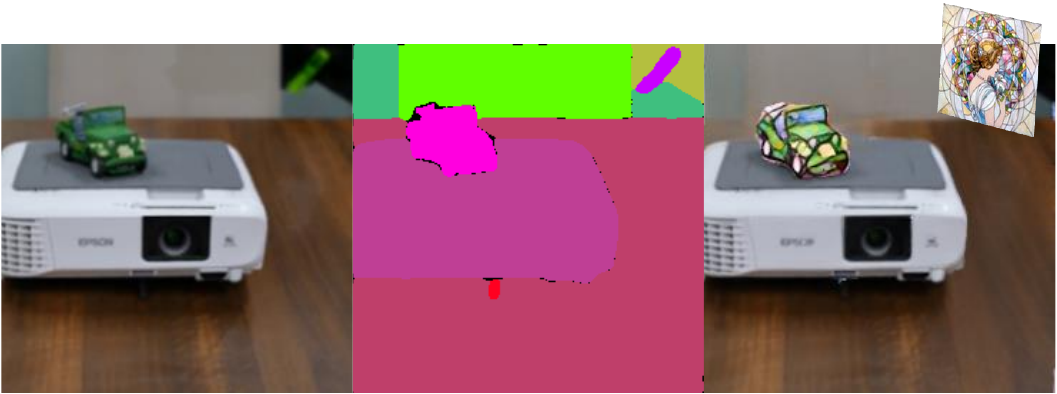}
  \caption{Spectral scene editing: The segmented scene at 450nm (middle) is used to perform a semantic style-transfer on the full spectra (left). The semantic stylized scene (right) has been generated using by applying a style transfer on the multi-view maps (full-spectra) and then in-painting the splats using the semantic object-ID in spectrum 450nm.}
  \label{fig:scene_editing}
\end{figure}

\begin{table*}[htb!]
\label{tab:spectraNeRFSyntheticDataset}
\centering
\caption{Quantitative Comparisons (PSNR / SSIM / LPIPS) on Spectral NeRF \textbf{synthetic} Dataset \cite{spectralnerf}}
\begin{tabular}{|c|c|c|c|c|c|c|c|c|}
\hline
\multirow{2}{*}{Method} & \multicolumn{6}{c|}{Spectral NeRF Synthetic Dataset\cite{spectralnerf}} & \multirow{2}{*}{Average} \\
\cline{2-7}
& kitchen & Living room & Digger & Spaceship & Vintage car & Cartoon knight & \\
\hline
& \multicolumn{6}{c|}{PSNR $\uparrow$} & \\
\hline
NeRF\cite{mildenhall2020nerf} & 34.583 & 33.172 & 30.658 & 30.126 & 33.478 & 34.485 & 32.400 \\
Mip-NeRF\cite{barron2021mipnerf} & - & - & 33.301 & 31.495 & 33.883 & \cellcolor{yellow!40}35.102 & \cellcolor{yellow!40}33.945 \\
Aug-NeRF\cite{chen2022augnerf} & 34.480 & 32.540 & 31.538 & 30.929 & 33.639 & 33.908 & 32.677 \\
SpectralNeRF\cite{spectralnerf} & \cellcolor{yellow!40}35.115 & \cellcolor{yellow!40}33.665 & \cellcolor{yellow!40}33.378 & \cellcolor{yellow!40}31.951 & \cellcolor{yellow!40}34.480 & 34.915 & 33.610 \\
\hline
Ours & \cellcolor{orange!40}37.035 & \cellcolor{orange!40}37.989 & \cellcolor{orange!40}40.218 & \cellcolor{orange!40}41.233 & \cellcolor{orange!40}42.636 & \cellcolor{orange!40}36.723 & \cellcolor{orange!40}38.456 \\
\hline\hline
& \multicolumn{6}{c|}{SSIM $\uparrow$} & \\
\hline
NeRF\cite{mildenhall2020nerf} & 0.8943 & 0.9929 & 0.9187 & 0.9358 & 0.7958 & 0.9273 & 0.9123 \\
Mip-NeRF\cite{barron2021mipnerf} & - & - & 0.9290 & 0.9475 & 0.8166 & \cellcolor{yellow!40}0.9572 & 0.9126 \\
Aug-NeRF\cite{chen2022augnerf} & 0.9026 & 0.9649 & 0.9248 & 0.9402 & 0.8002 & 0.9287 & 0.9163 \\
SpectralNeRF\cite{spectralnerf} & \cellcolor{yellow!40}0.9117 & \cellcolor{orange!40}0.9931 & \cellcolor{yellow!40}0.9357 & \cellcolor{yellow!40}0.9482 & \cellcolor{yellow!40}0.8169 & \cellcolor{orange!40}0.9573 & \cellcolor{yellow!40}0.9349 \\
\hline
Ours & \cellcolor{orange!40}0.9747 & \cellcolor{yellow!40}0.9733 & \cellcolor{orange!40}0.9923 & \cellcolor{orange!40}0.9951 & \cellcolor{orange!40}0.9893 & \cellcolor{yellow!40}0.9572 & \cellcolor{orange!40}0.9801 \\
\hline\hline
& \multicolumn{6}{c|}{LPIPS $\downarrow$} & \\
\hline
NeRF\cite{mildenhall2020nerf} & 0.1650 & 0.0578 & 0.0413 & 0.0275 & \cellcolor{yellow!40}0.1319 & 0.1545 & \cellcolor{yellow!40}0.0722 \\
Mip-NeRF\cite{barron2021mipnerf} & - & - & 0.0435 & 0.0535 & 0.1747 & 0.1526 & 0.1061 \\
Aug-NeRF\cite{chen2022augnerf} & \cellcolor{yellow!40}0.1603 & 0.0706 & 0.0341 & 0.0389 & 0.1536 & 0.1705 & 0.0973 \\
SpectralNeRF\cite{spectralnerf} & 0.1637 & \cellcolor{orange!40}0.0479 & \cellcolor{yellow!40}0.0259 & \cellcolor{yellow!40}0.0250 & 0.1499 & \cellcolor{yellow!40}0.1510 & 0.0733 \\
\hline
Ours & \cellcolor{orange!40}0.0739 & \cellcolor{yellow!40}0.0525 & \cellcolor{orange!40}0.0109 & \cellcolor{orange!40}0.0084 & \cellcolor{orange!40}0.0527 & \cellcolor{orange!40}0.0741 & \cellcolor{orange!40}0.0438 \\
\hline
\end{tabular}
\end{table*}

\subsection{Spectral scene editing}
Our framework extends scene editing techniques, such as Gaussian Grouping~\cite{gaussian_grouping}, into the spectral domain, unlocking a wide range of possibilities. By leveraging the semantic information present in any of the spectrum maps, we can achieve object deletion, in-painting, and style-transfer. Figure~\ref{fig:scene_editing} illustrates the utilization of segmentation maps obtained from the 450 nm spectrum for the stylization of the splats across the full spectra.

To accomplish this, we transfer the style to the multi-view full spectra maps and perform object in-painting through a fine-tuning of the splats, similar to Gaussian grouping~\cite{gaussian_grouping},  using the new ground truth (multi-view semantic stylized maps). The significance of this capability is particularly evident in fields like cultural heritage, where the retrieval of color information from a specific spectral band enables the accurate restoration of missing color details throughout the full-spectrum. By leveraging these advancements, we can enhance various applications and open up new avenues for exploration. 

\section{Experiments}
To demonstrate the potential of our approach, we provide both quantitative and qualitative evaluations with comparisons to baseline techniques.

\subsection{Baseline techniques used for comparison}
The techniques used as a reference in the scope of the evaluation include several state-of-the-art variants of Neural Radiance Fields (NeRF) (i.e., NeRF~\cite{mildenhall2020nerf}, MIP-NeRF~\cite{barron2021mipnerf}, Aug-NeRF~\cite{chen2022augnerf}, Ref-NeRF~\cite{verbin2022refnerf}) (which considers appearance parameters) and Gaussian splatting (i.e., Gaussian splatting without special reflectance modeling~\cite{kerbl3Dgaussians} and Gaussian Shader that specifically models reflectance~\cite{jiang2023gaussianshader}) as well as the respective extensions of such modern scene representation approaches to the spectral domain (i.e., SpectralNeRF~\cite{spectralnerf} and Cross-spectral NeRF~\cite{poggi2022xnerf}).


\begin{figure}[htb!]
\centering
  \includegraphics[width=\linewidth]{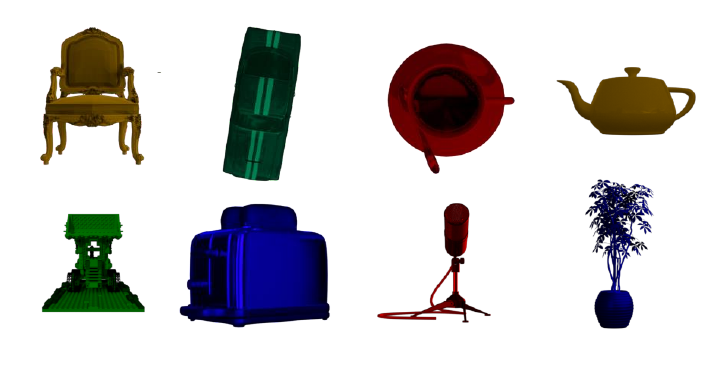}
  \caption{Snapshot of the different scenes in the Spectral NeRF synthetic and Spectral shiny Blender datasets}
  \label{fig:datasets_snapshot}
\end{figure}
\subsection{Datasets}
\label{sec:datasets}
\begin{figure*}[htb!]
\centering
  \includegraphics[width=\linewidth]{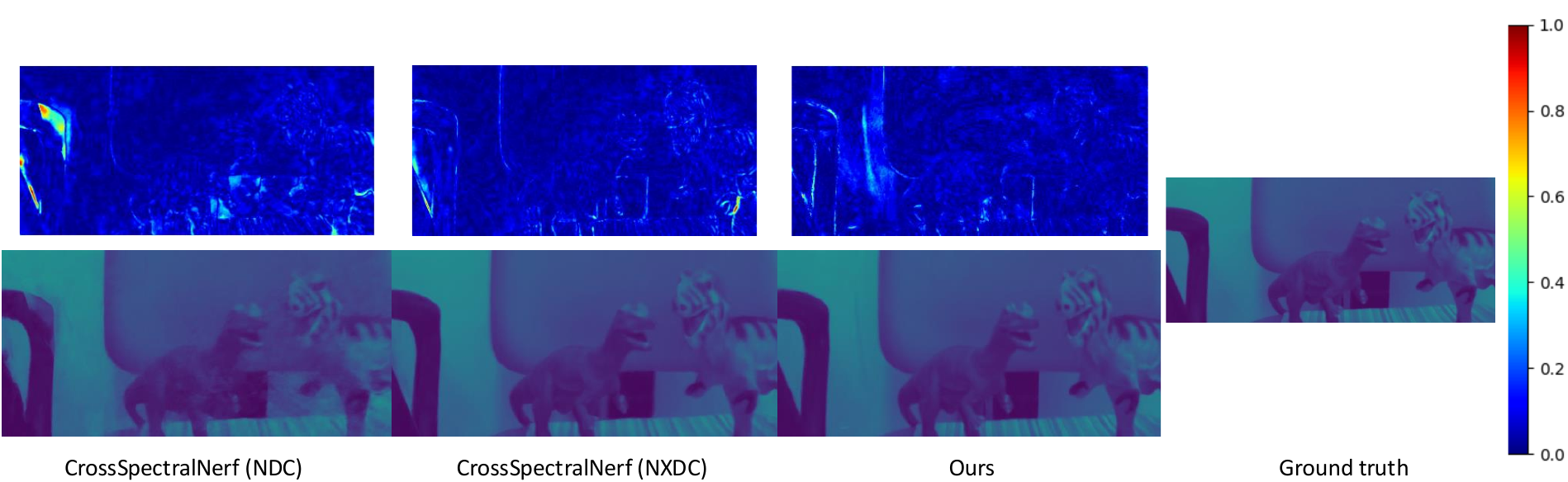}
  \caption{Qualitative comparison of CrossSpectralNerf \cite{poggi2022xnerf} with our method with the dino dataset.}
  \label{fig:compare_cross_dino}
\end{figure*}

For the comparison with  SpectralNeRF, we use both synthetic and real-world multi-spectral videos \cite{spectralnerf}. The poses for the digger, spaceship, and vintage car models were estimated using DUSt3R~\cite{dust3r2023} since reconstruction failed with COLMAP~\cite{schoenberger2016sfm}. For the remaining scene videos (kitchen, living room, projector, and dragon doll), COLMAP was used to generate the poses.

To demonstrate the adaptability of our method in handling cross-spectral data (infrared and multi-spectral), we conducted a comparative analysis using the cross-spectral NeRF dataset~\cite{poggi2022xnerf}. We created the ground truth full spectrum image from the cross-spectral spectrum-maps. For this, we averaged the images from all spectra and applied the colormaps \emph{viridis} and \emph{magma} for the multi-spectral and infrared dataset respectively, similar to the approach used in cross-spectral NeRF~\cite{poggi2022xnerf}.
To further validate that the spectral appearance estimation produces plausible results for different types of scenes (having also highly-reflective objects in the scene), we created a synthetic multi-spectral dataset from the shiny blender dataset \cite{verbin2022refnerf} and synthetic NeRF dataset \cite{mildenhall2020nerf} (see Figure \ref{fig:datasets_snapshot}). We generated this multi-spectral dataset using Mitsuba~\cite{Mitsuba3} for 5 bands from 460nm to 620nm similar to SpectralNeRF~\cite{spectralnerf}. We generated the data for the scenes where the shading model supported in Mitsuba corresponded to the shading model in Blender in order to get representative data. We utilized this dataset to conduct a comparative analysis of our method against state-of-the-art NeRF and Gaussian splatting techniques. The results are presented in Table~\ref{tab:spec_synthetic_nerf} and Table~\ref{tab:spectral_shiny_blender}.


\begin{table}[h]
\centering
\caption{Dataset Overview \\ *MS = Multispectral, *IR = Infrared}
\scalebox{0.56}{
\begin{tabular}{|>{\centering\arraybackslash}m{2.5cm}|>{\centering\arraybackslash}m{2.5cm}|>{\centering\arraybackslash}m{2.5cm}|>{\centering\arraybackslash}m{2.5cm}|>{\centering\arraybackslash}m{2.5cm}|}
\hline
\multicolumn{1}{|c|}{Dataset} & \multicolumn{1}{c|}{Scenes} & \multicolumn{1}{>{\centering\arraybackslash}p{2.5cm}|}{Number of multi-view images} & \multicolumn{1}{>{\centering\arraybackslash}p{2.5cm}|}{Number of iterations} & \multicolumn{1}{>{\centering\arraybackslash}p{2.5cm}|}{Number of spectral bands} \\
\hline
SpectralNeRF & 6 synthetic and 2 real-world (MS)* & 20 (Digger, Spaceship, Vintage car), 40 (cartoon knight) and 120 (all other scenes) & 40,000 (Digger, Spaceship, Vintage car), 30,000 (all other scenes) & 5 (Synthetic) and 8 (Real) \\
\hline
CrossSpectralNeRF & 16 real-world (MS + IR)* & 30 - 32 & 30,000 & 10 (MS) and 1(IR) \\
\hline
Spectral ShinyBlender& 5 synthetic (MS)* & 120 & 30,000 & 5 \\
\hline
Spectral SyntheticNeRF & 4 Synthetic (MS)* & 120 & 30,000 & 5 \\
\hline
\end{tabular}}
\label{tab:datasets}
\end{table}
\subsection{Implementation details}
The evaluations were conducted on an Nvidia RTX 3090 graphics card. In most scenes, we used a total of 30,000 iterations, except for the digger, spaceship, and vintage car scenes where we used 40,000 iterations. For the comparison to other methods, we used the results reported in their original publications.

\subsection{Quantitative analysis}

\begin{table}[h]
\label{tab:spec_nerf_real}
\centering
\caption{Quantitative Comparisons (PSNR / SSIM / LPIPS) on Spectral NeRF \textbf{real} Dataset\cite{spectralnerf}}
\scalebox{0.73}{
\begin{tabular}{|c|c|}
\hline
\multirow{2}{*}{Method} & \multicolumn{1}{c|}{Spectral NeRF \textbf{real} Dataset\cite{spectralnerf}} \\
\cline{2-2}
 & Projector \\
\hline
\hline
& \multicolumn{1}{c|}{PSNR $\uparrow$}\\
\hline
NeRF\cite{mildenhall2020nerf} &  28.9670 \\
Aug-NeRF\cite{chen2022augnerf}  & 30.0795 \\
SpectralNeRF\cite{spectralnerf}  &  \cellcolor{yellow!40}31.2535 \\
\hline
Ours   & \cellcolor{orange!40}35.8949 \\
\hline
\hline
& \multicolumn{1}{c|}{SSIM $\uparrow$}\\
\hline
NeRF\cite{mildenhall2020nerf} &  0.9429 \\
Aug-NeRF\cite{chen2022augnerf}  & \cellcolor{yellow!40}0.9573 \\
SpectralNeRF\cite{spectralnerf}  & 0.9449 \\
\hline
Ours  & \cellcolor{orange!40}0.9702 \\
\hline
\hline
& \multicolumn{1}{c|}{LPIPS $\downarrow$}\\
\hline
NeRF\cite{mildenhall2020nerf}  &  \cellcolor{yellow!40}0.0472 \\
Aug-NeRF\cite{chen2022augnerf}  & \cellcolor{orange!40}0.0354 \\
SpectralNeRF\cite{spectralnerf}  & 0.0605 \\
\hline
Ours  & 0.0882 \\
\hline
\end{tabular}}
\end{table}

\begin{table}[h]
\label{tab:cross_spec_nerf_real}
\centering
\caption{Quantitative comparison (PSNR / SSIM) with the cross-spectral NeRF dataset \cite{poggi2022xnerf}}
\scalebox{0.83}{
\begin{tabular}{|c|c|c|c|c|c|}
\hline
\multicolumn{4}{|c|}{Configuration} & \multicolumn{2}{c|}{Avg.} \\
\cline{1-4} \cline{5-6}
Model & Train & NXDC & Test & PSNR & SSIM \\
\hline
NeRF & MS & - & MS & 33.53 & 0.917 \\
X-NeRF & RGB+MS & $\times$ & MS & 31.96 & 0.897 \\
X-NeRF & RGB+MS & $\checkmark$ & MS & \cellcolor{yellow!40}33.87 & \cellcolor{yellow!40}0.918 \\
\hline\hline
NeRF & MS & - & MS & 33.53 & 0.917 \\
X-NeRF & RGB+MS+IR & $\times$ & MS & 30.87 & 0.870 \\
X-NeRF & RGB+MS+IR & $\checkmark$ & MS & 33.53 & 0.914 \\
\hline\hline
Ours & RGB+MS & - & MS & \cellcolor{orange!40}35.17 & \cellcolor{orange!40}0.962 \\
\hline\hline
NeRF & IR & IR & - & \cellcolor{orange!40}33.26 & \cellcolor{yellow!40}0.897 \\
X-NeRF &  RGB+MS+IR & IR & $\times$ & 31.60 & 0.869 \\
X-NeRF &  RGB+MS+IR & IR & $\checkmark$ & 32.44 & 0.879 \\
\hline\hline
Ours & RGB+IR & - & IR & \cellcolor{yellow!40}33.19 &  \cellcolor{orange!40}0.952 \\
\hline
\end{tabular}}
\end{table}
Quantitative analysis was performed on all datasets mentioned in Section \ref{sec:datasets} and overview of the number of scenes, multi-view images and number of iterations for which each scene was trained is presented in Table \ref{tab:datasets}. We compute the PSNR \cite{psnr}, SSIM \cite{nilsson2020understandingssim} and LPIPS \cite{lpips} for all camera-views and report average the average result.  
The orange in the tables represents the best result and yellow represents the second best results. 
\begin{figure*}[htb!]
\centering
  \includegraphics[width=\linewidth]{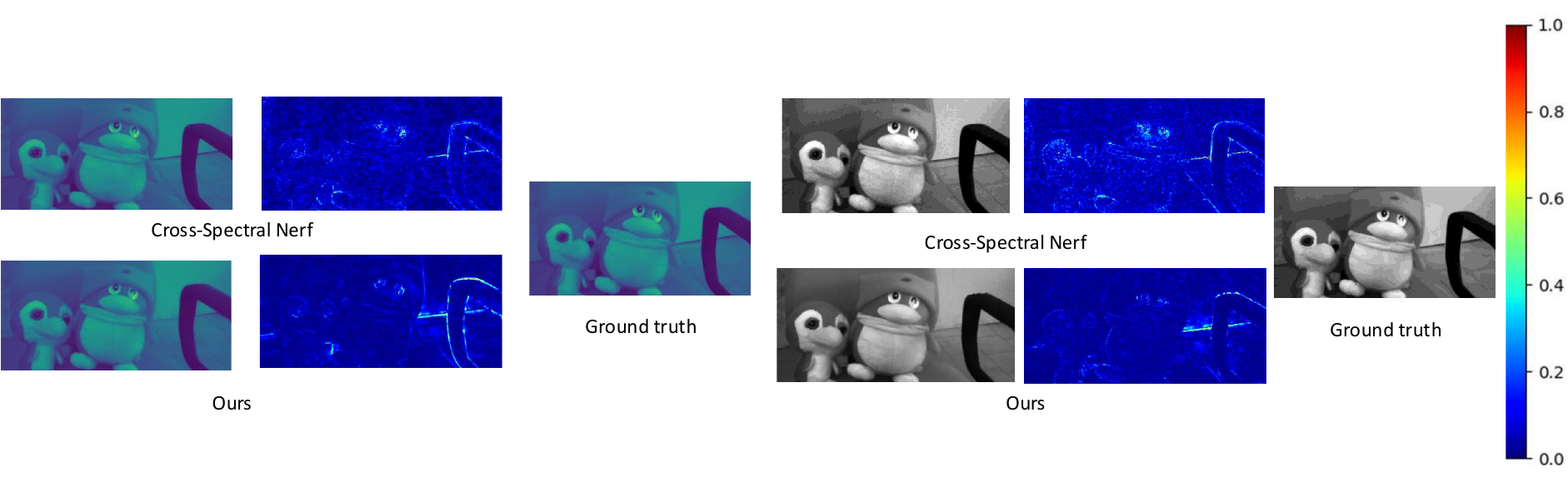}
  \caption{Qualitative comparison of CrossSpectralNerf \cite{poggi2022xnerf} with our method with the Penguin dataset. The comparison shows the average of the 10 spectra colored with colormap viridis (left) and one such spectra (right)}
  \label{fig:compare_cross_penguin}
\end{figure*}
\subsubsection{Comparison with radiance-field-based spectral methods}
The quantitative analysis shows that our method overall outperforms the existing spectral methods~\cite{poggi2022xnerf, spectralnerf} for both multi-spectral and cross-spectral data. The results presented in Table~\ref{tab:spectraNeRFSyntheticDataset} indicate that our method outperforms SpectralNeRF in most scenes and on average for the synthetic dataset. Additionally, our analysis, as shown in Table~\ref{tab:spec_nerf_real}, reveals that our method also surpasses SpectralNeRF when applied to the real-world dataset. It is important to note that due to the unavailability of all datasets and test views from the original paper, our evaluation was limited to only one real-world dataset (see Table \ref{tab:spec_nerf_real}) for the SpectralNeRF method. However, we also compare our method based on the Cross-spectral NeRF dataset which contains only real-world scenes. Here, our method clearly performs better for all scenes (multi-spectral and infrared datasets) as presented in Table~\ref{tab:cross_spec_nerf_real}. This shows that our method produces plausible results with real-world scenes and outperforms state-of-the-art spectral methods.

\begin{table}[h]
\label{tab:spectral_shiny_blender}
\centering
\caption{Quantitative Comparisons (PSNR / SSIM / LPIPS) on Spectral shiny Blender dataset}
\scalebox{0.63}{
\begin{tabular}{ccccccc}
\hline
\multirow{2}{*}{Method} & \multicolumn{5}{c}{Spectral Shiny Blender Dataset} \\
\cline{2-7}
& Car & Helmet & Teapot & Toaster & Coffee & Avg. \\
\hline
& \multicolumn{5}{c}{PSNR $\uparrow$}\\
\hline
NVDiffRec\cite{Munkberg_2022_CVPR} & 27.98 & 26.97 & 40.44 & 24.31 & 30.74 & 28.70 \\
NVDiffMC\cite{hasselgren2022nvdiffrecmc} & 25.93 & 26.27 & 38.44 & 22.18 & 29.60 & 28.88 \\
Ref-NeRF\cite{verbin2022refnerf} & \cellcolor{yellow!40}30.41 & 29.92 & 45.19 & 25.29 & \cellcolor{yellow!40}33.99 & \cellcolor{yellow!40}32.32 \\
NeRO\cite{liu2023nero} & 25.53 & 29.20 & 38.70 & \cellcolor{orange!40}26.46 & 28.89 & 29.84 \\
ENVIDR\cite{liang2023envidr} & 28.46 & \cellcolor{yellow!40}32.73 & 41.59 & 26.11 & 29.48 & 32.88 \\
Guassian Splatting\cite{kerbl3Dgaussians} & 27.24 & 28.32 & \cellcolor{yellow!40}45.68 & 20.99 & 32.32 & 30.37 \\
Gaussian shader\cite{jiang2023gaussianshader} & 27.90 & 28.32 & \cellcolor{orange!40}45.86 & \cellcolor{yellow!40}26.21 & 32.39 & 31.94 \\
\hline
Ours & \cellcolor{orange!40}30.37 & \cellcolor{orange!40}36.39 & 44.42 &  24.82 & \cellcolor{orange!40}36.62 & \cellcolor{orange!40}34.524 \\
\hline\hline
& \multicolumn{5}{c}{SSIM $\uparrow$}\\
\hline
NVDiffRec\cite{Munkberg_2022_CVPR}& \cellcolor{yellow!40}0.963 & 0.951 & \cellcolor{orange!40}0.996 & 0.928 & \cellcolor{orange!40}0.973 & 0.945 \\
NVDiffMC\cite{hasselgren2022nvdiffrecmc} & 0.940 & 0.940 & \cellcolor{yellow!40}0.995 & 0.886 & 0.965 & 0.944 \\
Ref-NeRF\cite{verbin2022refnerf} & 0.949 & 0.955 & \cellcolor{yellow!40}0.995 & 0.910 & \cellcolor{yellow!40}0.972 & 0.956 \\
NeRO\cite{liu2023nero} & 0.949 & \cellcolor{orange!40}0.971 & \cellcolor{yellow!40}0.995 & 0.929 & 0.956 & \cellcolor{yellow!40}0.962 \\
ENVIDR\cite{liang2023envidr} & 0.961 & 0.980 & \cellcolor{orange!40}0.996 & \cellcolor{yellow!40}0.939 & 0.949 & \cellcolor{orange!40}0.969 \\
Guassian Splatting\cite{kerbl3Dgaussians} & 0.930 & 0.951 & \cellcolor{orange!40}0.996 & 0.895 & 0.971 & 0.947 \\
Gaussian shader\cite{jiang2023gaussianshader} & 0.931 & 0.950 & \cellcolor{orange!40}0.996 & 0.929 & 0.971 & 0.957 \\
\hline
Ours & \cellcolor{orange!40}0.970 & \cellcolor{yellow!40}0.970 & 0.992 & \cellcolor{orange!40}0.942 & \cellcolor{orange!40}0.973 & \cellcolor{orange!40}0.969 \\
\hline\hline
& \multicolumn{5}{c}{LPIPS $\downarrow$}\\
\hline
NVDiffRec\cite{Munkberg_2022_CVPR}& \cellcolor{orange!40}0.045 & 0.118 & \cellcolor{yellow!40}0.011 & 0.169 & \cellcolor{yellow!40}0.076 & 0.119 \\
NVDiffMC\cite{hasselgren2022nvdiffrecmc} & 0.077 & 0.157 & 0.014 & 0.225 & 0.097 & 0.147 \\
Ref-NeRF\cite{verbin2022refnerf} & 0.051 & 0.087 & 0.013 & 0.118 & 0.082 & 0.109 \\
NeRO\cite{liu2023nero} & 0.074 & \cellcolor{yellow!40}0.050 & 0.012 & 0.089 & 0.110 & 0.072 \\
ENVIDR\cite{liang2023envidr} & \cellcolor{yellow!40}0.049 & 0.051 & \cellcolor{yellow!40}0.011 & \cellcolor{yellow!40}0.116 & 0.139 & 0.072 \\
Guassian Splatting\cite{kerbl3Dgaussians} & 0.047 & 0.079 & \cellcolor{orange!40}0.007 & 0.126 & 0.078 & 0.083 \\
Gaussian Shader\cite{jiang2023gaussianshader} & \cellcolor{orange!40}0.045 & 0.076 & \cellcolor{orange!40}0.007 & \cellcolor{orange!40}0.079 & 0.078 & \cellcolor{yellow!40}0.068 \\
\hline
Ours & 0.049 & \cellcolor{orange!40}0.043 & 0.026 & \cellcolor{orange!40}0.079 & \cellcolor{orange!40}0.068 & \cellcolor{orange!40}0.053\\ 
\hline\hline
\end{tabular}}
\end{table}

\subsubsection{Comparison with non-spectral radiance-field-based methods}
To demonstrate that our method produces plausible results compared to existing state-of-the-art Gaussian splatting methods~\cite{jiang2023gaussianshader,kerbl3Dgaussians}, we conducted a comparison using spectral datasets created from both the NeRF synthetic dataset and the shiny Blender dataset, as described in Section~\ref{sec:datasets}. The analysis reveals that our method consistently outperforms existing methods on average for the Shiny Blender dataset, as shown in Table~\ref{tab:spectral_shiny_blender}. This indicates that extending Gaussian splatting to the spectral domain improves the accuracy of reflectance estimation, particularly for shiny objects. Additionally, our method performs quite well on the synthetic NeRF dataset, as evidenced by the average PSNR and SSIM values in Table~\ref{tab:spec_synthetic_nerf}.


\begin{table}[htb!]
\label{tab:spec_synthetic_nerf}
\centering
\caption{Quantitative Comparisons (PSNR / SSIM / LPIPS) on Spectral Synthetic NeRF dataset}
\scalebox{0.70}{
\begin{tabular}{ccccccc}
\hline
\multirow{2}{*}{Method} & \multicolumn{5}{c}{Spectral Synthetic NeRF Dataset} \\
\cline{2-6}
& Chair & Lego & Mic & Ficus & Avg. \\
\hline
& \multicolumn{5}{c}{PSNR $\uparrow$}\\
\hline
NeRF\cite{mildenhall2020nerf} & 33.00 & 32.54 & 32.91 & 30.13 & 32.64 \\
VolSDF\cite{yariv2021volume} & 30.57 & 29.46 & 30.53 & 22.91 & 28.87 \\
Ref-NeRF\cite{verbin2022refnerf} & 33.98 & 35.10 & 33.65 & 28.74 & 32.11 \\
ENVIDR\citet{liang2023envidr} & 31.22 & 29.55 & 32.17 & 26.60 & 29.88 \\
Gaussian Splatting\cite{kerbl3Dgaussians} & 35.82 & \cellcolor{yellow!40}35.69 & \cellcolor{yellow!40}35.34 & 34.83 & 35.17 \\
Gaussian Shader\cite{jiang2023gaussianshader} & \cellcolor{yellow!40}35.83 & \cellcolor{orange!40}35.87 & 35.23 & \cellcolor{yellow!40}34.97 & \cellcolor{yellow!40}35.22 \\
\hline
Ours &  \cellcolor{orange!40}38.93 &  34.26 & \cellcolor{orange!40}36.80 & \cellcolor{orange!40}36.57 & \cellcolor{orange!40}36.39 \\
\hline\hline
& \multicolumn{5}{c}{SSIM $\uparrow$}\\
\hline
NeRF\cite{mildenhall2020nerf} & 0.967 & 0.961 & 0.980 & 0.964 & 0.968 \\
VolSDF\cite{yariv2021volume} & 0.949 & 0.951 & 0.969 & 0.929 & 0.949 \\
Ref-NeRF\cite{verbin2022refnerf} & 0.974 & 0.975 & 0.983 & 0.954 & 0.971 \\
ENVIDR\cite{liang2023envidr} & 0.976 & 0.961 & 0.984 & \cellcolor{yellow!40}0.987 & 0.977 \\
Gaussian Splatting\cite{kerbl3Dgaussians} & \cellcolor{yellow!40}0.987 & \cellcolor{orange!40}0.983 & \cellcolor{orange!40}0.991 & \cellcolor{yellow!40}0.987 & \cellcolor{orange!40}0.987 \\
Gaussian Shader\cite{jiang2023gaussianshader} & \cellcolor{yellow!40}0.987 & \cellcolor{orange!40}0.983 & \cellcolor{orange!40}0.991 & 0.985 & \cellcolor{yellow!40}0.986 \\
\hline
Ours & \cellcolor{orange!40}0.990 &  \cellcolor{yellow!40}0.977 &  \cellcolor{yellow!40}0.990 & \cellcolor{orange!40}0.994 & \cellcolor{orange!40}0.987 \\
\hline\hline
& \multicolumn{5}{c}{LPIPS $\downarrow$}\\
\hline
NeRF\cite{mildenhall2020nerf} & 0.046 & 0.050 & 0.028 & 0.044 & 0.042 \\
VolSDF\cite{yariv2021volume} & 0.056 & 0.054 & 0.191 & 0.068 &  0.092 \\
Ref-NeRF\cite{verbin2022refnerf} & 0.029 & 0.025 & 0.018 & 0.056 & 0.032 \\
ENVIDR\cite{liang2023envidr} & 0.031 & 0.054 & 0.021 & \cellcolor{yellow!40}0.010 & 0.029 \\
Gaussian Splatting\cite{kerbl3Dgaussians} & \cellcolor{orange!40}0.012 & \cellcolor{yellow!40}0.016 & \cellcolor{orange!40}0.006 & 0.012 & \cellcolor{yellow!40}0.012 \\
Gaussian Shader\cite{jiang2023gaussianshader} & \cellcolor{orange!40}0.012 & \cellcolor{orange!40}0.014 & \cellcolor{orange!40}0.006 & 0.013 &  \cellcolor{orange!40}0.011 \\
\hline
Ours & \cellcolor{yellow!40}0.017 & 0.031 &  \cellcolor{yellow!40}0.014 & \cellcolor{orange!40}0.006 &  0.017 \\
\hline\hline
\end{tabular}}
\end{table}

\subsection{Qualitative analysis}
We conducted a qualitative comparison between our method and the Cross-spectral NeRF~\cite{poggi2022xnerf} using the dino and penguin datasets. The results, shown in Figure~\ref{fig:compare_cross_dino} for the dino dataset and Figure~\ref{fig:compare_cross_penguin} for the penguin dataset, highlight the superior performance of our method in reconstructing scene appearance. In particular, Figure~\ref{fig:compare_cross_penguin} demonstrates the accurate rendering of specular effects in the eyes of the penguin, showcasing the effectiveness of our approach. Additionally, Figure~\ref{fig:compare_cross_dino} reveals that our method produces better reflectance reconstruction, as evidenced by the shading effects on the surface of the dino.

\begin{figure*}[htb!]
\centering
  \includegraphics[width=\linewidth]{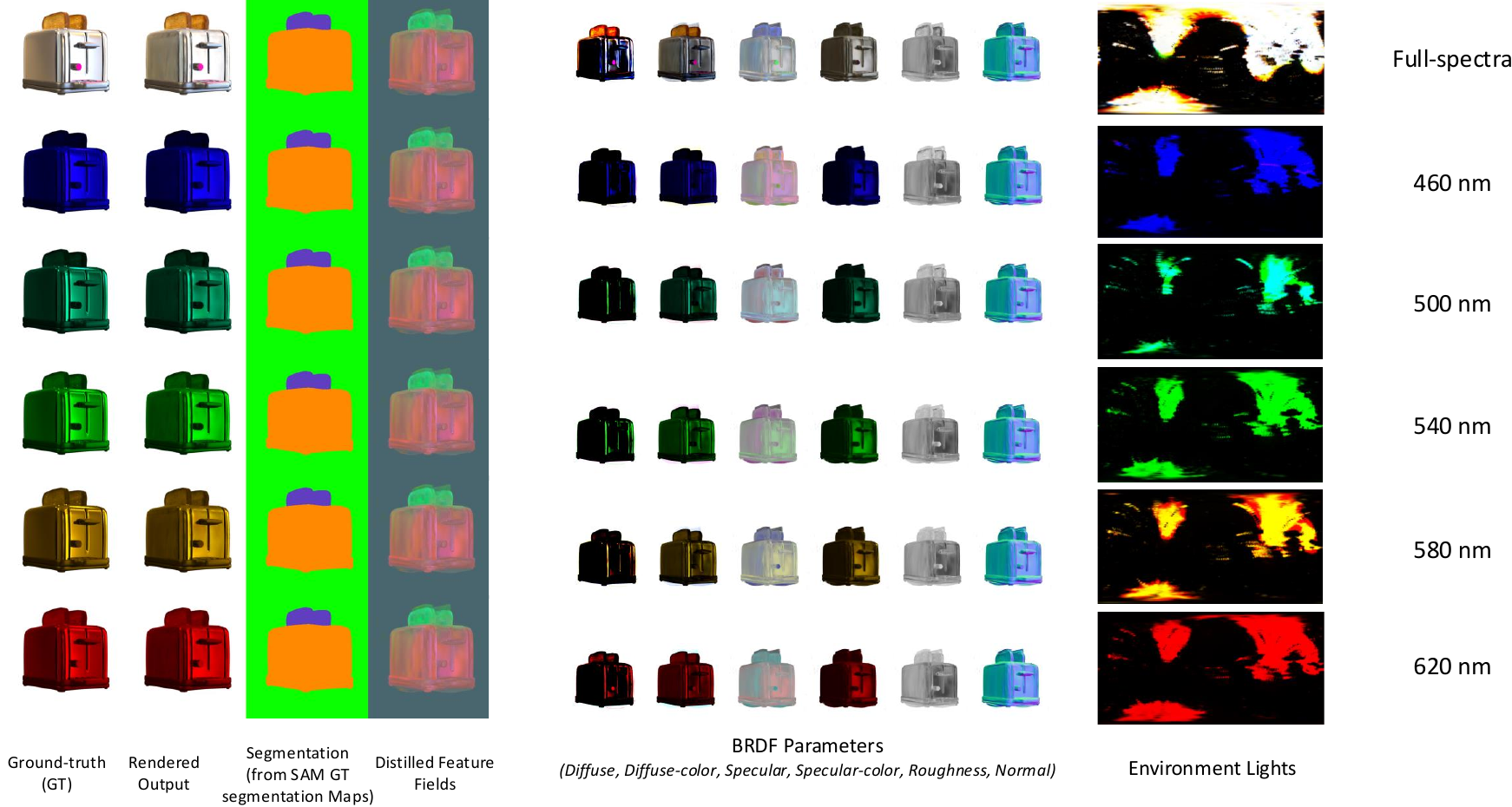}
  \caption{Qualitative analysis (Rendering of the toaster scene using our framework from multi-spectral data)}
  \label{fig:qualitative_full_spectra}
\end{figure*}

As depicted in Figure~\ref{fig:qualitative_full_spectra}, our framework successfully estimates the lighting and BRDF parameters within the individual spectra, while also providing segmented object IDs. This showcases the effectiveness and accuracy of our framework in capturing and analyzing the desired parameters for the given scene.

\subsection{Ablation study}

In this section, we conduct ablations by eliminating the warm-up iterations that we introduced to enhance reflectance and light estimations in the scene through the inclusion of appropriate priors from other spectra. For this, we use three real-world scenes: dragon doll (from the SpectralNeRF dataset~\cite{spectralnerf}), orange, and tech scenes (from the Cross-SpectralNeRF dataset~\cite{poggi2022xnerf}). The dragon doll scene has 8 bands, while the orange and tech scenes have 10 bands.
\begin{table}[htb!]
\label{tab:ablation}
\centering
\caption{
%
%
Ablation studies comparing full-spectra reconstructions with (a) no prior initialization from other spectra and (b) learnable parameters in full spectra initialized with priors from other spectra after a warm-up iteration show that the latter approach yields better results.}
%
\scalebox{0.99}{
\begin{tabular}{cccccc}
\hline
 & Dragon doll & Orange & Tech & hall4 & Avg. \\
\hline
& \multicolumn{4}{c}{PSNR $\uparrow$}\\
\hline
(a) & 36.55 & 42.98 & 40.17 & 40.99 & 40.17 \\
\hline
(b) & \cellcolor{orange!40}38.52 & \cellcolor{orange!40}44.13 & \cellcolor{orange!40}40.73 &  \cellcolor{orange!40}42.14 & \cellcolor{orange!40}41.63 \\
\hline\hline
& \multicolumn{4}{c}{SSIM $\uparrow$}\\
\hline
(a) & 0.972 & 0.992 & 0.986 & 0.989 & 0.985 \\
(b) & \cellcolor{orange!40}0.980 & \cellcolor{orange!40}0.994 & \cellcolor{orange!40}0.987 & \cellcolor{orange!40}0.991 & \cellcolor{orange!40}0.988 \\
\hline\hline
& \multicolumn{4}{c}{LPIPS $\downarrow$}\\
\hline
(a) & 0.047 & 0.017 & \cellcolor{orange!40}0.045 & 0.018 & 0.031 \\
(b) & \cellcolor{orange!40}0.029 & \cellcolor{orange!40}0.013 & 0.051 & \cellcolor{orange!40}0.017& \cellcolor{orange!40}0.027\\ 
\hline\hline
\end{tabular}}
\end{table}

\begin{figure*}[htb!]
\centering
  \includegraphics[width=\linewidth]{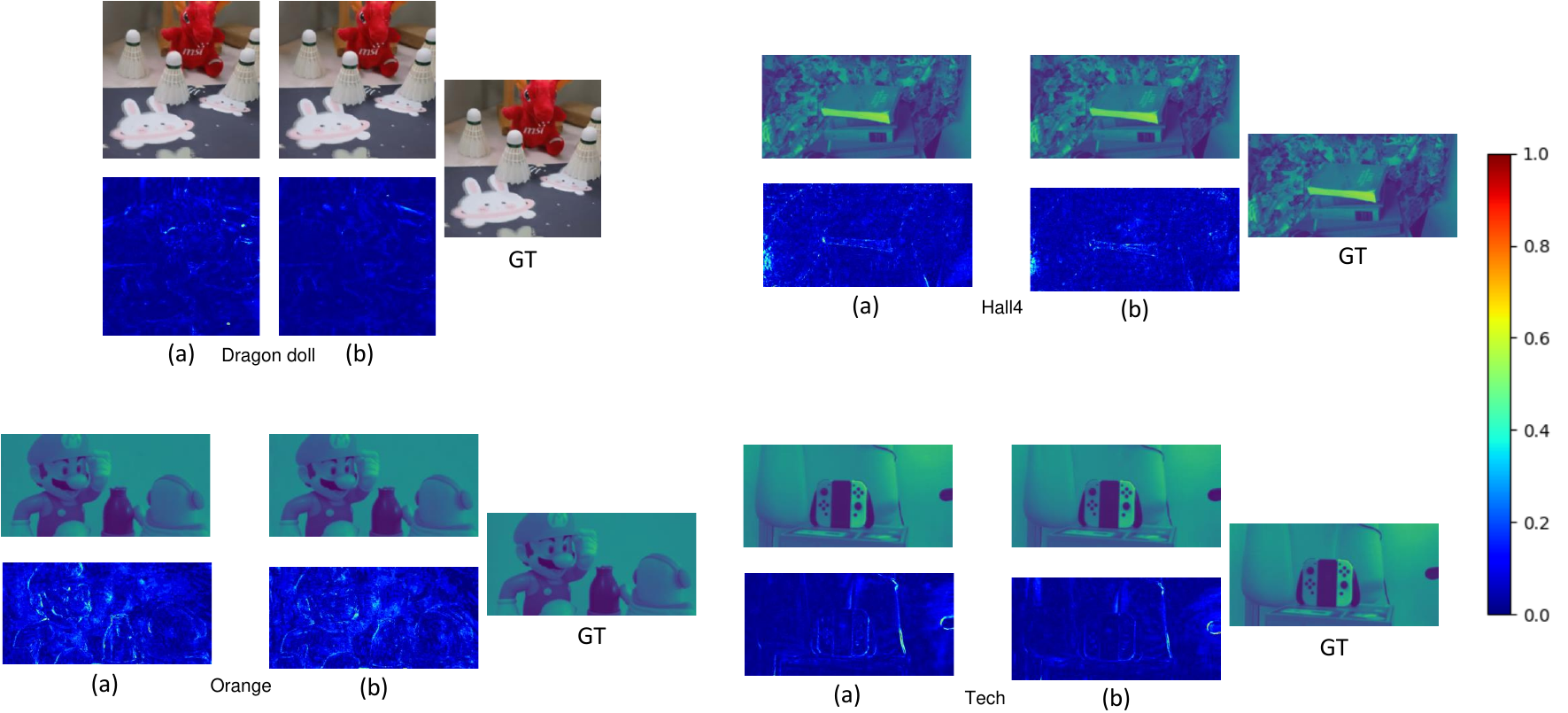}
  \caption{Ablation studies were conducted to assess the differences with ground truth for scenes dragon-doll \cite{spectralnerf}, hall4, orange, and tech \cite{poggi2022xnerf}. The models were trained under two conditions: (a) without initialization of full-spectra model parameters from other spectra, and (b) with initialization of full-spectra model parameters using the average of common model parameters from other spectra.}
  \label{fig:qualitative_ablation}
\end{figure*}

To evaluate the impact of including priors from different spectra, we conducted a comprehensive analysis, encompassing both quantitative measurements (see Table ~\ref{tab:ablation}) and qualitative observations (see Figure~\ref{fig:qualitative_ablation}), after initializing the common model parameters with the average of all other spectra following a warm-up phase of 1000 iterations.

\paragraph{Quantitative analysis:}
The results presented in Table~\ref{tab:ablation} clearly indicate that incorporating information from other spectra leads to improved average performance metrics for the rendered output across different real-world scenes. The higher average values achieved regarding PSNR and SSIM and the lower LPIPS values demonstrate enhancements when utilizing additional spectral information, highlighting the effectiveness of this approach in improving rendering quality and material asset estimation.

\paragraph{Qualitative analysis:}
In addition to the quantitative analysis, we conducted a qualitative assessment by comparing the rendered outputs with the ground truth for the aforementioned scenes. The results reveal noticeable improvements in capturing finer details, such as the edges of the shuttlecock in the Dragon doll scene, as well as enhanced reconstruction of objects like the speaker in the tech scene (see Figure~\ref{fig:qualitative_ablation}). These findings further reinforce the effectiveness of incorporating information from other spectra in achieving more accurate and detailed rendering results.

\subsection{Limitations}
%
While the presented framework offers promising capabilities, it is important to acknowledge its limitations. One such limitation is the requirement for spectrum-maps to be co-registered, which can be a complex and time-intensive process. Moreover, as the resolution of images increases and more spectra are incorporated, the training time escalates significantly. To overcome these challenges, future research can explore the integration of alternative deep learning algorithms that support end-to-end training specifically for co-registering maps. Additionally, improving the encoding methods to efficiently accommodate a larger number of spectra would enhance the framework's capabilities.
%

%
Another limitation to consider is that the shading model currently used in the framework is fixed. However, the framework can be modified to have a flexible number of learnable parameters based on the shading model. This would allow users to configure the framework to their specific needs and enable more customized and adaptable shading models. By addressing these limitations, the framework can be made more practical and effective, enabling seamless co-registration, support for an expanded range of spectra, reduced training time for high-resolution images, and user-configurable shading models.
\section{Conclusion}
We presented 3D Spectral Gaussian Splatting, a cross-spectral rendering framework that utilizes 3D Gaussian Splatting to generate realistic and semantically meaningful splats from registered multi-view spectrum and segmentation maps. This framework enhances scene representation by incorporating multiple spectra, providing valuable insights into material properties and segmentation. Additionally, the paper introduces an improved physically-based rendering approach for Gaussian splats, enabling accurate estimation of reflectance and lights per spectra, resulting in enhanced realism. Furthermore, the paper showcases the potential of spectral scene understanding for precise scene editing techniques such as style transfer, in-painting, and removal.The contributions of this work address challenges in multi-spectral scene representation, rendering, and editing, opening up new possibilities for diverse applications.

Future work can focus on improving the accuracy of lighting and reflectance estimation in the proposed framework.
%
While we demonstrated our approach to outperform other recent, spectral learning-based scene representations ~\cite{poggi2022xnerf,spectralnerf} for different scenes, the evaluation of its potential for high-precision scanning with costly devices like the TAC7~\cite{merzbach2017high}, that allow capturing lots of photographs under controlled light-view conditions, might be interesting as well. There might be a chance that our learning-based spectral scene representation offers advantages over the parametric models used as a default option for the TAC7 due to the flexibility of the learnable models.
%
%
Additionally, the utilization of spectral data, which has not been used in learning-based scene representation techniques like NeRFs or 3D Gaussian Splatting with a careful reflectance modeling so far, can open up new possibilities for achieving better results in this field. Additionally, integrating a registration process into the pipeline would allow for end-to-end training of non-co-registered spectrum maps, which is common with many spectral cameras. Exploring these areas can lead to better results and expand the possibilities of research in this field and open new opportunities for several applications where spectral characteristics are of great importance.

\section*{Acknowledgement}
The work presented in this paper has been partially funded by the European Commission during the project PERCEIVE under grant agreement 101061157.

 \bibliographystyle{elsarticle-num} 
 \bibliography{spectral_gaussian}
\end{document}